\documentclass[conference]{IEEEtran}
\IEEEoverridecommandlockouts
\usepackage{cite}
\usepackage{amsmath,amssymb,amsfonts}
\usepackage{algorithmic}
\usepackage[ruled,vlined]{algorithm2e}
\usepackage{graphicx}
\usepackage{textcomp}
\usepackage{subfigure}
\usepackage{xcolor}
\usepackage{comment}
\usepackage{mathtools}  
\usepackage{amsfonts}  

\def\BibTeX{{\rm B\kern-.05em{\sc i\kern-.025em b}\kern-.08em
    T\kern-.1667em\lower.7ex\hbox{E}\kern-.125emX}}

\begin{document}

\title{ESOD:Edge-based Task Scheduling for Object Detection}

\author{\IEEEauthorblockN{Yihao Wang\IEEEauthorrefmark{2}, Ling Gao\IEEEauthorrefmark{2}\IEEEauthorrefmark{1},  Jie Ren\IEEEauthorrefmark{3}\IEEEauthorrefmark{1}, Rui Cao\IEEEauthorrefmark{2}, Hai Wang\IEEEauthorrefmark{2}, Jie Zheng\IEEEauthorrefmark{2}, Quanli Gao\IEEEauthorrefmark{3}}

\IEEEauthorblockA{
\IEEEauthorrefmark{2}Northwest University, China\\
\IEEEauthorrefmark{3}Shannxi Normal University, China, \\
\IEEEauthorrefmark{3}Xian Polytechnic University, China,\\
gl@nwu.edu.cn, renjie@snnu.edu.cn
\vspace{-10 mm} }

}

\maketitle

\begin{abstract}

Object Detection on the mobile system is a challenge in terms of everything. Nowadays, many object detection models have been designed, and most of them concentrate on precision. However, the computation burden of those models on mobile systems is unacceptable. Researchers have designed some lightweight networks for mobiles by sacrificing precision. We present a novel edge-based task scheduling framework for object detection (termed as ESOD) . In detail, we train a DNN model (termed as pre-model) to predict which object detection model to use for the coming task and offloads to which edge servers by physical characteristics of the image task (e.g., brightness, saturation). The results show that ESOD can reduce latency and energy consumption by an average of 22.13\% and 29.60\% and improve the mAP to 45.8(with 0.9 mAP better), respectively, compared with the SOTA DETR model.

\end{abstract}


\section{Introduction}
\subsection{Background}

Object detection has always been a challenge for the mobile terminal in many aspects. In practical applications such as unmanned driving, aerospace, virtual reality and other scenarios, it is necessary to quickly and accurately locate the category and location of the object. Since Yann LeCun first applied LeNet\cite{b1} to the MNIST task in 1994 and achieved great improvement, a large number of deep neural network-based models have occupied a dominant position in computer vision. Although the precision of these models is increasing, the complexity of computation is also increasing more rapidly than which. These dramatic increases in computing costs have made it more and more difficult to deploy those models on the mobile system. Most of the existing models can complete the inference in a short time on the PC, but this is based on the powerful computing power of the graphics processing unit (GPU). However, there is almost no such computing power support on the mobile, so there is no way to perform the  interference  task of object detection on the mobile terminal using those models. In addition, even if the object detection model is successfully implemented, the huge energy consumption and time overhead caused by the calculation is terrible.

In this paper, we present \textbf{ESOD}(Edge-Based Dynamic Scheduling Strategy for Deep Learning Object Detection Task) which off-loads various tasks on different edge servers by using a pre-classify model to model the physical characteristics of the image and the precision of the prediction model. With ESOD, better detection precision can be obtained on mobile devices at a low computing cost. To sum up, our contributions can be summarized as follows: First, we propose a task scheduling strategy based on edge computing; secondly, we propose an index to balance model precision and model cost. Finally, we consider the limitations of multiple edge computing platforms.

\subsection{Related Work}
\paragraph{Object Detection Model}
To solve the exhaustive drawbacks of the sliding window, Alex et al.\cite{b2} applied the convolutional neural network to computer vision first and proposed AlexNet. To further improve the precision, Simonyan et al.\cite{b3} proposed the VGG network by increasing the number of convolutional layers and reducing the size of the convolution kernel, which brought object detection into the deep models. He et al.\cite{b4} proposed ResNet through residual connection technology, which solves the problem that the object detection network is difficult to superimpose to the deep layer, which makes the detection network deeper. Based on this, Girshick et al.\cite{b5}\cite{b6}\cite{b7}\cite{b8} proposed the R-CNN series methods, which first proposed a two-stage object detection model. However, because of the slow speed of the two-stage model, one-stage models such as YOLO\cite{b9} and SSD\cite{b10} are gradually invented. Aiming to further improve the detection precision of small objects, FPN\cite{b11} technology came into being, and with FPN the object detection network becomes deeper. The manual design of the Ancher mechanism of these object detection models has been a problem that people want to improve for a long time, Anchor Free models such as FCOS\cite{b12}, Center Mask\cite{b13} and other models have also been gradually designed. In recent years, due to the success of using the Transformer mechanism to replace the CNN mechanism model in the natural language processing, Carion et al.\cite{b14} has also proposed a Transformer-based object detection model named DETR for the shortcomings of the CNN-based model being limited to local information. However, these models without exception increase the complexity of the model and the computational requirements of the model, and precision improvement obtained is relatively slight.

\begin{figure}[htbp]
	\center
	\subfigure[Task 1]{\label{fig:1a}
	\includegraphics[width=0.45\linewidth]{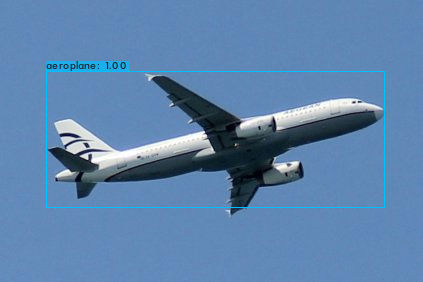}}
	\subfigure[Task 2]{\label{fig:1b}
	\includegraphics[width=0.45\linewidth]{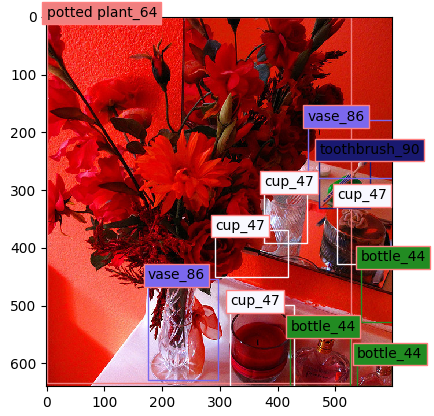}}
\caption{Different Task}
\label{fig:moti_demo}
\end{figure}

\begin{figure*}[htbp]
	\centering
	\subfigure[Energy consumption]{\includegraphics[width=0.37\textwidth]{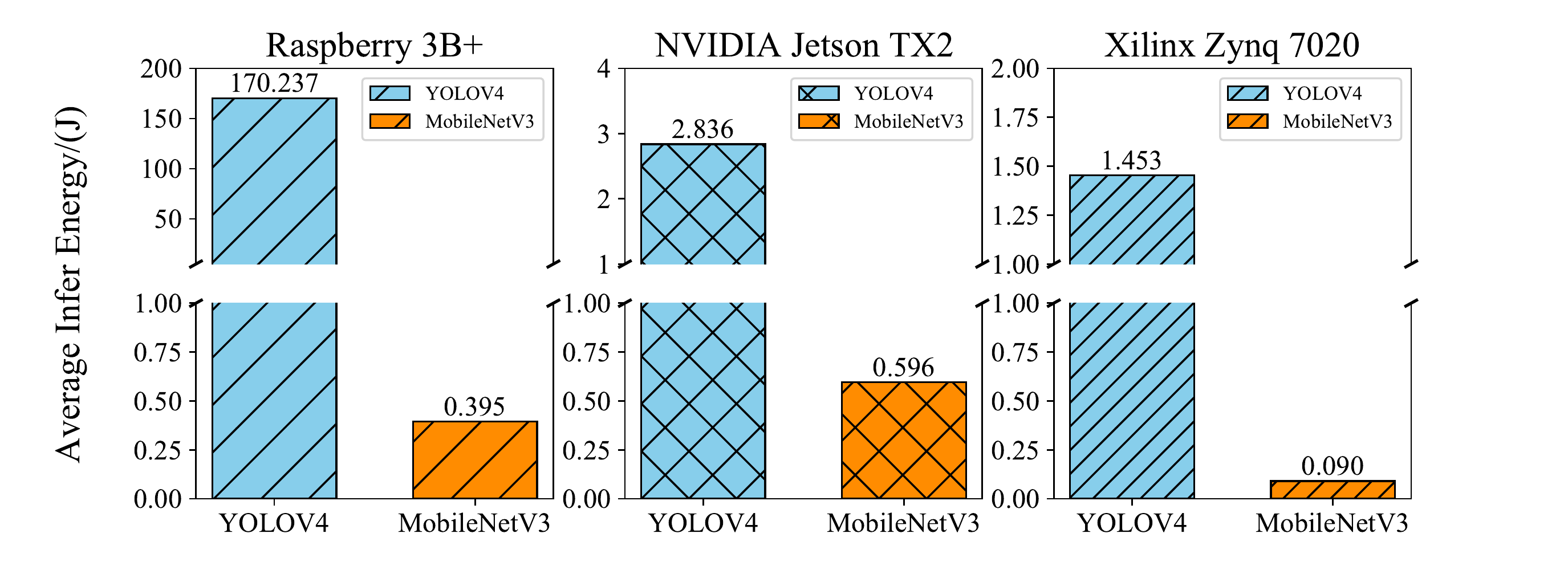}}
    \subfigure[Inference time]{\includegraphics[width=0.37\textwidth]{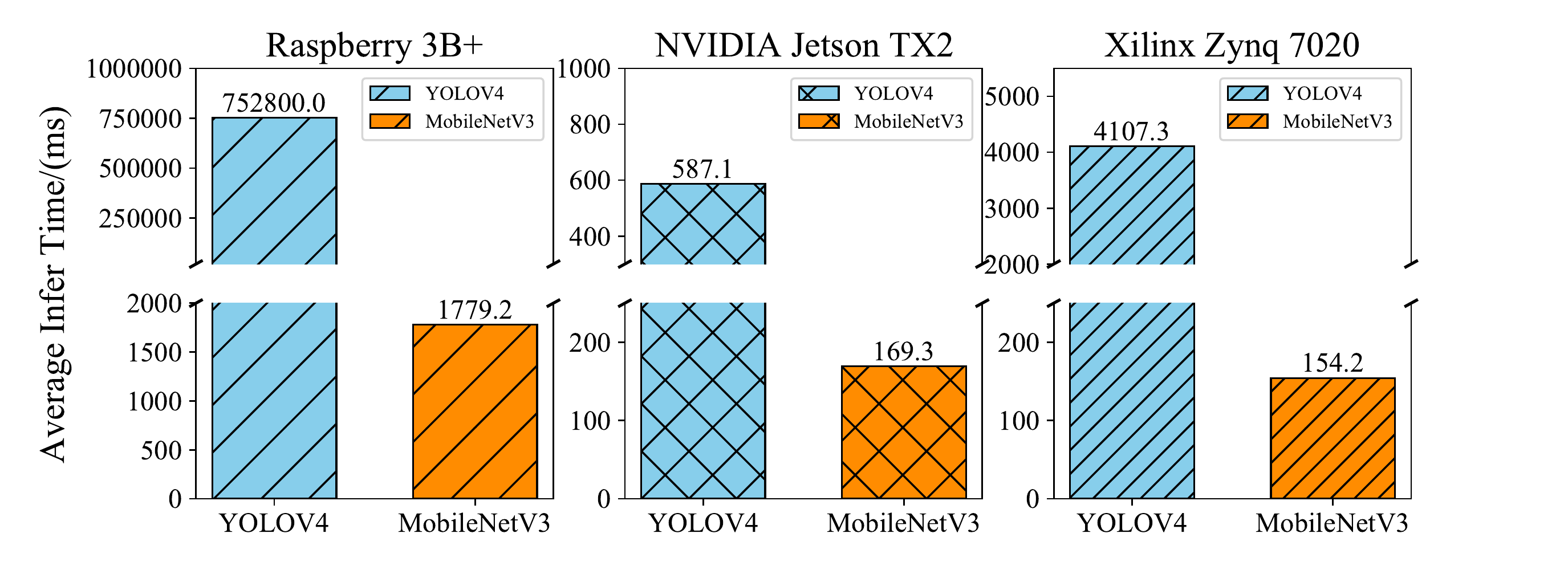}}
    \subfigure[IoU]{\includegraphics[width=0.2\textwidth]{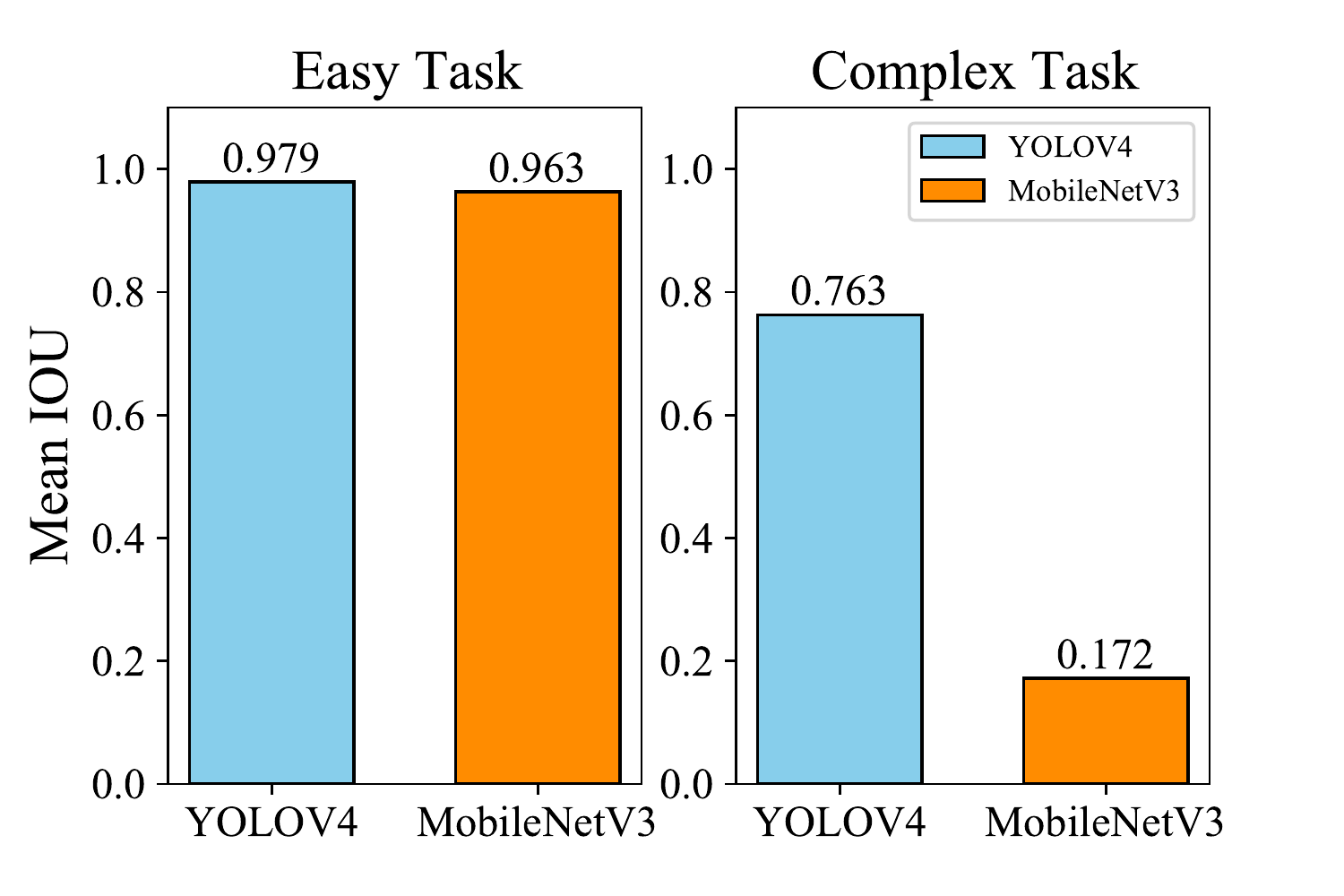}}

    \caption{Average energy consumption (a), inference time (b) and IoU (c) when applying two SOTA object detection models on image 1 and image 2.}
    \label{fig:moti_results}
\end{figure*}

\paragraph{Lightweight Network}
The heavy computation cost of CNN-based detection models on mobile hinders the development of these models so that Andrew G et al.\cite{b15} proposed a deep separable convolution. By improving the traditional convolution of VGG into a deep separable convolution, MobileNetV1 achieves the same precision as SSD with 10x fewer computation costs. Aiming at the shortcomings of MobileV1 training that it is difficult to converge, by combining the connection method of ResNet, Sandler et al.\cite{b16} proposed inverted convolution and improved MobileNetV2. By proposing the combination of integer quantization activation function and NAS technology, Andrew G et al.\cite{b17} proposed MobileNetV3 by further reducing the number of layers of MobileNetV2. Zhang et al.\cite{b18} improved the 1x1 convolution in MobileNetV2 to a Shuffle operation, and further proposed an object detection model ShuffleNet with a smaller computational complexity. Through research on the different characteristics of hardware devices, Ma et al.\cite{b19} improved ShuffleNet and proposed ShuffleNetV2, and further designed a model with a lower computational load on mobile devices. However, although these models can reduce the computational overhead on the mobile terminal, they only make the deep target detection technology possible on the mobile terminal. In terms of effects, there is a big gap between the network model designed for the mobile terminal and the traditional model.

Then, whether the precision of the traditional model can be obtained, and the energy consumption can be reduced on the mobile terminal has become a problem studied in this paper.

\section{Motivation}
As a motivating example, we consider performing object detection on two SOTA models, YoloV4\cite{b22} and MobileNetV3-Small\cite{b17}. Our evaluation devices are three representative embedded platforms, RaspberryPi 3B+ (as the mobile device), NVIDIA Jetson TX2 (as the edge server), and FPGA-accelerated Xilinx Zynq-7020 (as the edge server). All these models are built upon TensorFlow v2.3.0 and have been pre-trained by using the Microsoft COCO 2017 dataset on the server.

Figure \ref{fig:moti_results} presents the inference time, energy consumption, and IoU of two models when performing object detection on images as shown in Figure \ref{fig:moti_demo}. As we can see that image 1 is quite a straightforward task, both MobileNetV3-Small and YOLOV4 provide the correct results, and the IoU is above 0.75 (0.75 is the strict detection metric used in the object detection task). While MobileNetV3-Small takes, on average, 0.36 J and 700.9 ms on image 1, which reduces 83.18\% energy and 99.72 \% inference latency than YoloV4. Clearly, for image 1, MobileNetV3-Small is good enough, and there is no need to use a more advanced (and costly) model for inference. Besides, it is worth noting that the performance of the three platforms is contrasting. YoloV4 achieves the fastest inference speed on TX2, while MobileNetV3-Small on Zynq 7020 performs best. In addition, when energy consumption is the priority, we would like to choose Zynq 7020 to perform object detection, which gives an average of 0.7715 J on two tasks. This experiment shows that choosing a suitable model and platform is vital for object detection task offloading.

\begin{figure}[htbp]
\centering
\includegraphics[scale=0.3]{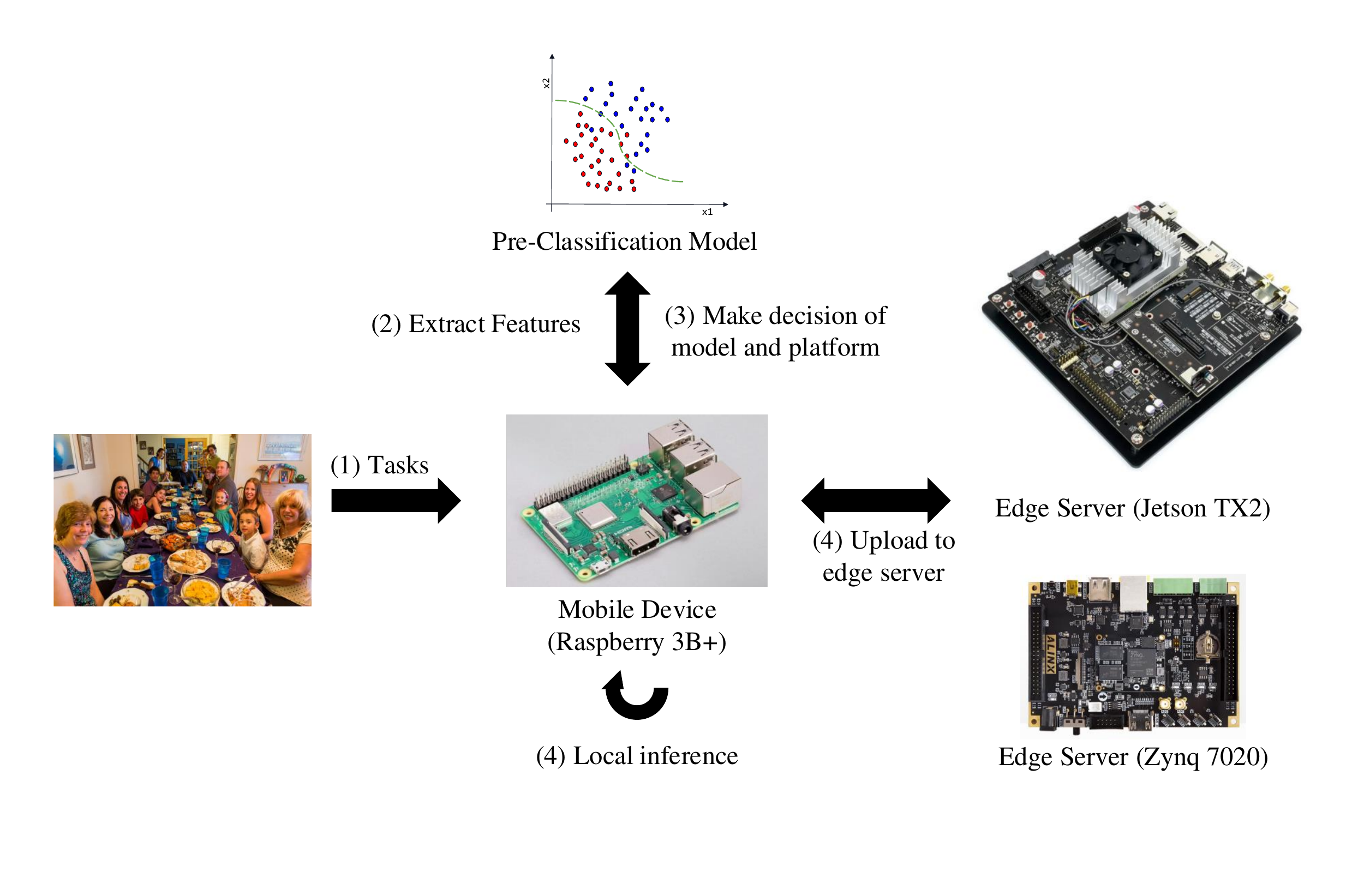}
\caption{Overview of our approach}
\label{fig-arch}
\end{figure}

%

\begin{table}
	\begin{center}
		\caption{Inference Time (ms)}
		\label{table6}
		\begin{tabular}{lccc} 
			\textbf{Model} & \textbf{Raspberry 3B+} & \textbf{TX2} & \textbf{Zynq 7020}  \\
			\hline
			YOLOv4 & 752800 & 587.1 & 4107.3 \\
			MobileNetV3 & 1779.2 & 169.3 & 154.2 \\
		\end{tabular}
	\end{center}
\end{table}

\begin{table}
	\begin{center}
		\caption{Energy Consumption (J)}
		\label{table-moti-energy}
		\begin{tabular}{lccc} 
			\textbf{Model} & \textbf{Raspberry 3B+} & \textbf{TX2} & \textbf{Zynq 7020}  \\
			\hline
			YOLOv4 & 170.237 & 2.84 & 1.45 \\
			MobileNetV3 & 0.395 & 0.596 & 0.09 \\
		\end{tabular}
	\end{center}
\end{table}

\begin{table}
	\begin{center}
		\caption{Mean IoU}
		\label{table-moti-iou}
		\begin{tabular}{lccc} 
			\textbf{Model} & \textbf{Easy Task} & \textbf{Complex Task} \\
			\hline
			YOLOv4 & 0.979 & 0.763 \\
			MobileNetV3 & 0.963 & 0.172 \\
		\end{tabular}
	\end{center}
\end{table}

\section{Our Approach}

\subsection{Overview}

To solve the various problems mentioned above and obtain a trade-off between precision and cost, motivated by \cite{b24,b25,b26,b27}, we proposed ESOD, an adaptive scheme to determine which deep model to use for a given task. ESOD provides a local response with low latency and low cost by deploying a lightweight object detection model on mobile devices while deploying a high-precision but high-cost deep learning model for detection tasks in complex scenarios on the edge server to ensure the precision of tasks. This process not only offers detection precision but also reduces the overhead (such as time and energy consumption) caused by inference as much as possible. Taking the object detection task as an example, the workflow of ESOD is shown in the figure \ref{fig-arch}:

\begin{itemize}
	\item [1)]
	Mobile application input image as a object detection task;
	\item [2)]
	The typical features representing the complexity of the image are extracted and normalized;
	\item [3)]
		Feed the processed features into the pre-classification model to predict the object detection model suitable for the current task;
	\item [4)]
	The selected model is used to decide whether to offload tasks to the edge server;
	\item [5)]
	Execute the prediction and return the result.
\end{itemize}

\subsection{Formulation}
\label{section32}

Let $Model_1,Model_2,...,Model_i,...,Model_n$ be different models; $BBox^j(Model_i)$ represents the detection result of the $Model_i$ on the sample. 
$Loss^j_i$ is the difference between the predicted result and the Ground Truth, $E(Model_i, Platform_k)$ represents the inference energy for this task using model i on platform k. $T_{i,k}$ represents the reasoning time of model i on platform k. $\hat T$ represents the threshold of inference time, $\hat E$ represents the threshold of energy consumption, and $\hat Loss$ represents the threshold of the loss value. Let $(Model_i, picture_j,platform_k)$ be the precision of a certain task $picture_j$ under different platform k and model i. Therefore, for a given task j, it is necessary to find a suitable target detection model on a suitable edge server according to the complexity of the task while meeting the minimum requirements of time, energy consumption, and precision. Based on this, the optimization problem abstracted in this paper is shown in (formula \ref {eq1}).

\begin{equation}
\begin{split}
\label{eq1}
max_{i=0,k=0}^{i=n,k=m}(Model_i,picture_j,platform_k) , \\
s.t. 
    \left\{
        \begin{array}{ll}
            Loss^j_i  \textless  \hat {Loss} \\
            T_{i,k}  \textless   \hat T  \\
		E(Model_i,Platform_k) \textless \hat E
        \end{array}
    \right.
\end{split}
\end{equation}

Motivated by the successful use of bipartite graph $Loss$ in DETR, we use a similar $Loss$ to that of DERT. However, unlike DETR, it is uncertain that the model used in this paper produces the prediction box that matches exactly like DETR is generated. Therefore, we add the miscalculated bbox confidence loss $Loss_{conf}$ and the uncalculated bbox area loss $Loss_{area}$ to the $Loss$ in this paper. Moreover, we use the $L_2$ loss to replace the $L_1$ loss used by $Loss_{box}$ in DETR.With the new loss, the $Loss$ used in this paper is shown in (formula \ref {eq2}) , So the $Loss$ in this paper is composed of the matching box loss $Loss_{box}(b_i,\hat {b_i})$ generated by bipartite graph matching, and the area loss of the unmatched real frame $Loss_{area}(b_i ,\hat {b_i})$, the confidence loss of the unmatched error box $Loss_{conf}(b_i,\hat {b_i})$ consists of these three parts. 

\begin{equation}
\label{eq2}
Loss(b_i,\hat {b_i}) = Loss_{box}(b_i,\hat {b_i}) + Loss_{conf}(b_i,\hat {b_i}) + \\ 
Loss_{area}(b_i,\hat {b_i})
\end{equation}

The three parts of the loss of the formula \ref {eq2} are calculated by the formula \ref {eq3}, the formula \ref {eq4}, and the formula \ref {eq5}. 

\begin{equation}
\label{eq3}
Loss_{box}(b_i,\hat {b_i}) = \lambda_{iou} L_{iou}(b_i,\hat {b_i})+\lambda_{L_2}\Vert b_i - \hat {b_i} \Vert
\end{equation}

The formula \ref {eq3} is used to calculate the loss between matching boxes, $\lambda_{iou},\lambda_{L_2} \in \mathbb{R}$, and the Hungarian algorithm is used to generate $b_i$ and $\ hat {b_i}$ match between, 

\begin{equation}
\label{eq4}
Loss_{conf}(b_i,\hat {b_i}) = 2^c - 1
\end{equation}
The formula \ref {eq4} is used to calculate the confidence loss of the unmatched prediction box, and $c$ is the confidence of the unmatched bbox box. 

\begin{equation}
\label{eq5}
Loss_{area}(b_i,\hat {b_i}) = \frac {\Vert {b_i} \Vert} {\Vert mean(\hat {b_i}) \Vert}
\end{equation}
The formula \ref {eq5} is used to calculate the area loss of the unmatched real frame, ${\Vert mean(\hat {b_i}) \Vert}$ is the average of the size of all real target frames, ${ \Vert {b_i} \Vert}$ is the size of the unmatched real frame. 

In order to facilitate us to solve this complex optimization problem, we further simplifies the optimization problem abstracted above.

\begin{equation}
\label{eq6}
\begin{split}
Score_{i,j,k}=f^1(\alpha ,T_{i,k}) + f^2(\beta ,E_{i,k}) + f^3(\gamma ,Loss^j_i) \\
s.t. 
	\left\{
       \begin{array}{ll}
		\alpha , \beta , \gamma \geq 0 \\
		\alpha + \beta + \gamma = 1
		\end{array}
	\right.
\end{split}
\end{equation}

As shown in formula \ref{eq6}, in this paper, the total time of inference and transmission, the total energy consumption of inference and transmission, and the difference value of the inference result are processed by the fusion processing of formula \ref{eq6}, and the i-th picture is obtained on the j-th model under the k-th platform through the formula \ref{eq6}. The total score of, where $\alpha,\beta,\gamma$ represents the cost factor that the current user expects to optimize. The closer the factor is to 1, the user overtime optimizes the consumption corresponding to this factor.

\subsection{model selection}
\subsubsection{model deployment}
\begin{table}[h!]
  \begin{center}
    \caption{model selection}
    \label{table2}
    \begin{tabular}{llll} 
      \textbf{Name} & \textbf{backbone}  & \textbf{mAP} \\
      \hline
      DETR & ResNet+Transformer & 44.9\\
      EfficientDet\cite{b4} & EfficientNet& 40.2 \\
      ShuffleNet & ResNet-D  & 25.4\\
      MobileNet & ResNet-D & 23.5\\
      YOLOV4\cite{b22} & DarkNet & 43.5\\
      CenterMask & VoVNetV2 & 43.1\\
      SparseRCNN\cite{b23} & ResNet & 42.8\\
      FCOS & ResNet & 42.1\\
      MaskRCNN & ResNet & 39.8\\
      RetinaNet\cite{b21} & ResNet & 35.1\\
    \end{tabular}
  \end{center}
\end{table}

Based on the issues mentioned above, we propose a dynamic scheduling method for high-energy-efficiency target detection tasks based on task complexity by combining the respective advantages of the lightweight target detection model and the complex target detection model. By deploying efficient models on the mobile system to provide low-overhead services and deploying complex models on the edge server to provide high-quality services, the scheduling method proposed in this paper can reduce energy consumption and inference time caused by calculations while maintaining precision.

We use the model deployment algorithm to determine the deployment plan of the object detection model in this paper. The model deployment algorithm needs to balance the relationship between the overall prediction performance mAP, prediction time, and prediction energy consumption of different models according to the weight list of different preferences set by the user $\alpha$ $\beta$ $\gamma$. First, get the mAP of each model on a data set, inference time, and inference energy consumption of each model on different platforms. Then use the formula \ref{eq6} and calculate the score of the model on different platforms according to $\alpha$ $\beta$ $\gamma$. Finally, for each model, the platform with the highest score of this model is selected as the deployment platform of the model.

After measuring the object detection effect mAP and inference time on public data set MS COCO validation set, the object detection model selected in this paper is as described in table \ref{table2}. In addition, to further consider the local inference situation, we additionally select object detection algorithms MobileNetV3 and ShuffleNetV2 designed for mobile are used as alternatives for local inference. The backbone network, mAP, and theoretical calculations of these models are shown in table \ref{table2}.

\subsubsection{scheduling strategy}
\begin{algorithm}
\caption{Model selection algorithm}
\label{alg2}
\LinesNumbered 
\KwIn{pair-list,Task,pre-classification model}
\KwOut{model,platform}
let $\boldsymbol {X} \Leftarrow$ features of Task\;
out $\Leftarrow$ label predicted by $\boldsymbol {X}$ using  pre-classification model\;
select model and corresponding platform\;
\end{algorithm}

Firstly, the feature set of the data $\mathbf{X}$and the Label set corresponding to each piece of data $\mathbf{Label}$ are first constructed. The features of data are composed of the image features described in table \ref {table1}. In this paper, the OpenCV tool is used to extract all the image features described. By comparing the average IOU of all the models selected in this paper in the picture, the model with the largest average IOU is selected as the corresponding label of this data. Secondly, after the feature set and label set are obtained, data sets are divided randomly by using Scikit-Learn to divide the data into a training set and test set, in which the test set takes up 10\%. Finally, the model is built on all the model parameters, the training is carried out on the training set, and the corresponding precision is evaluated on the test set, and the model with the highest precision is selected as the final pre-classification model.

Then, we need to train the pre-classification model by using the features shown in the table to construct the features and use the calculation method shown in the formula \ref{eq6} to obtain the model with the highest score as the label. After the pre-classification model is trained, the scheduling algorithm in this article is shown in the algorithm \ref{alg2}. Then we can use it to make predictions.

After training, it is necessary to deploy the pre-classification model on the mobile platform, and deploy various object detection algorithms determined by the equation \ref{eq6} on the edge server. Secondly, for the incoming task on the mobile system, the characteristics of the task are extracted and sent to the pre-classification algorithm for prediction. Finally, select the model output by the pre-classification algorithm and the corresponding edge server platform. According to this process, we construct a target detection scheduling algorithm based on task complexity.

\section{Expirement}

\subsection{features}

To evaluate the complexity of the picture task as reasonably and comprehensively as possible, we select the 30 image features described in the table \ref{table1} as the reference for describing image complexity information. The specific meanings of the image features selected in this article are described in the table \ref{table1}, which respectively includes statistical information of the image (such as the mean value of the color channel, etc.) and basic information about the object (such as the number of corners, edge points, etc.) and texture information of the image (such as contrast and uniformity in different directions). After selecting the features, we use these features as the input features of the pre-classification model. Through these features, we can construct a relationship between the image and a suitable model.

\begin{table}

  \begin{center}
    \caption{features}
    \label{table1}
    \begin{tabular}{ll} 
      \textbf{Feature} & \textbf{Description} \\
      \hline
      kpNum & Number of SIFT key points \\
      brightnessMean & Mean value of brightness \\
      brightnessRMS & Root mean square of brightness \\
      size & The amount of memory occupied \\
      cornerNum & Number of Harris corner points\\
      edgeNum & Number of Canny edge points \\
      contoursNum & The number of contours in the picture \\
      maxPointNum & The number of points in the longest contour\\
      area & The area enclosed by the longest contour \\
      arcLength & Arc length enclosed by the longest contour \\
      redMean & The color mean of the red channel \\
      greenMean & The color mean of the green channel \\
      blueMean & The color mean of the blue channel \\
      contrast$\left\{ 1 \sim 4 \right\}$ & Contrast in four directions \\
      homogeneity$\left\{ 1 \sim 4 \right\}$ & Homogeneity in four directions \\
      energy$\left\{ 1 \sim 4 \right\}$ & Energy in four directions \\
      correlation$\left\{ 1 \sim 4 \right\}$ & Correlation in four directions \\
      \end{tabular}
  \end{center}
\end{table}

\subsection{Details}

MLP pre-classification model: We use a multi-layer perceptron based on TensorFlow as the pre-classification model. The network structure of the MLP pre-classification model is a 9-layer fully connected layer (1 input layer, 7 hidden layers, 1 output layer). The activation function of the 1st to 8th layer is selected as the ReLu function, and the activation function of the 9th layer is selected as the softmax function. The training set and the test set use randomly divided data sets, the ratio of the test set is 0.2, and random scrambling is performed. Batch size is selected as 8, Adam with default parameters is selected as the optimizer, the learning rate is selected as 0.001, the loss function is used CategoricalCrossentropy is calculated, and the label is processed with one hot, and label smoothing with a parameter of 0.05 is performed, and the epoch is selected as 300. In the end, the pre-classification model in this paper can reach an accuracy of 84.06\% on the test set.

\section{Results}

\begin{figure}[htbp]
	\center
	\subfigure[The precision of different pre-classification models]{\label{fig:result:cost}
		\includegraphics[width=1.0\linewidth]{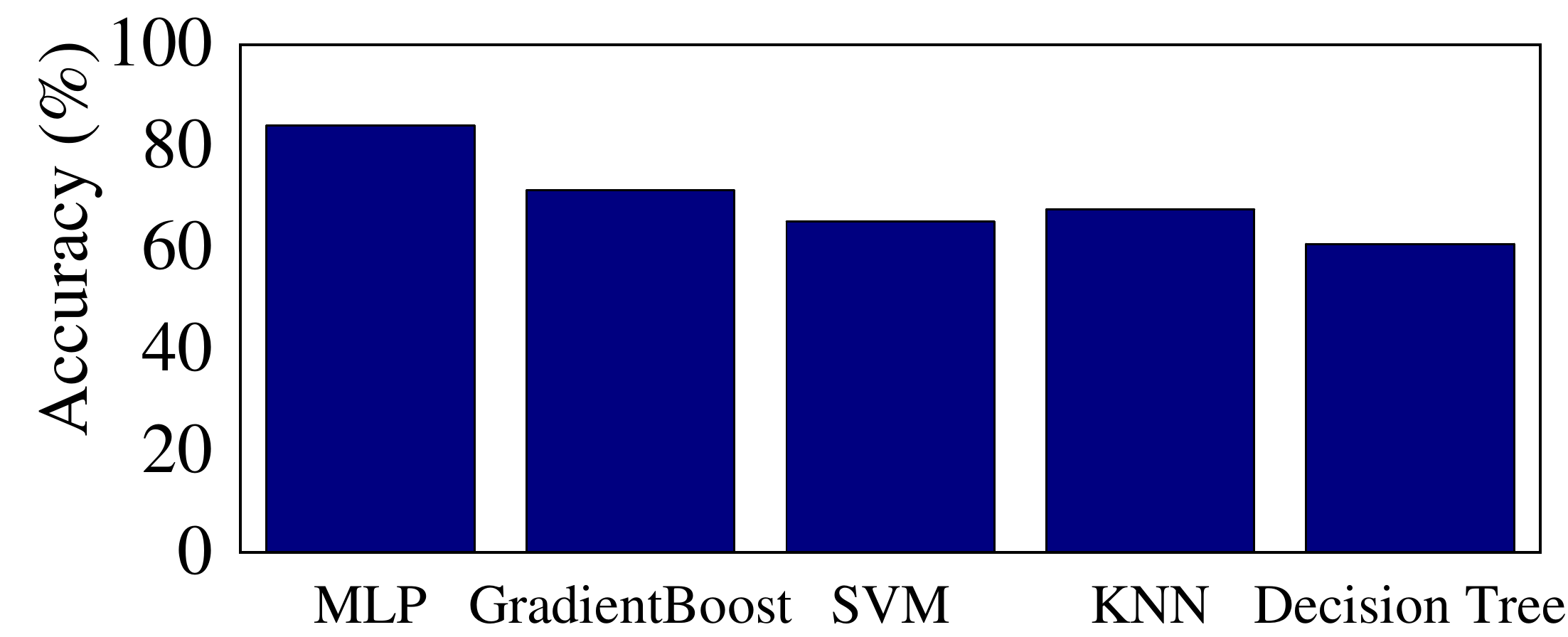}}
	\subfigure[The precision of the preclassification model under different optimization strategies]{\label{fig:result:acc}
		\includegraphics[width=1.0\linewidth]{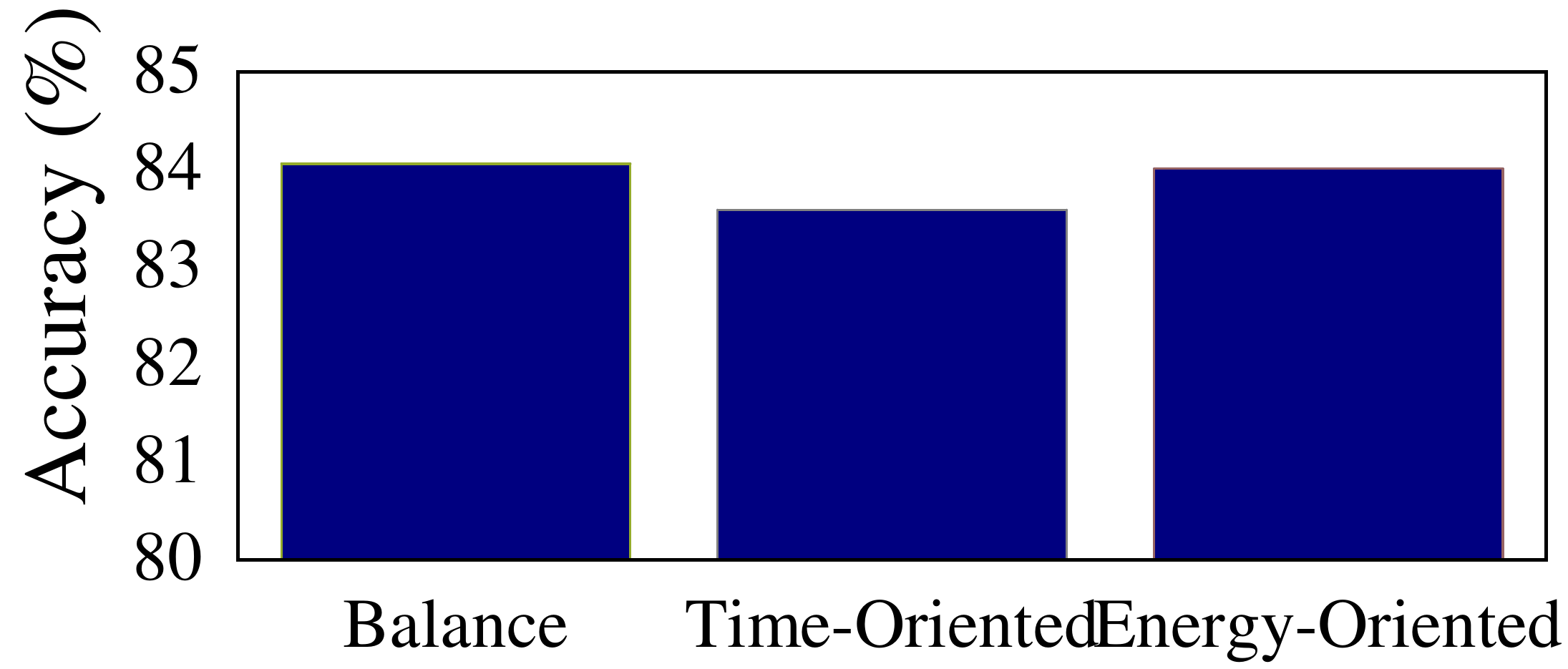}}
	\caption{The precision of different preclassification models (a) and different optimization strategies (b)}
	\label{fig:result}
\end{figure}

\subsection{precision}
Compared with DETR, the mAP after scheduling through the pre-classification model increased by 0.9, $AP_{75}$ increased by 4.0, and $AP_{M}$ increased by 2.7. Compared with YOLOV4, the pre-classified and scheduled model $AP_{50}$ Increased by 0.3, $AP_{S}$ increased by 29.6, but compared to DETR's effect on the Large target is much weaker, but compared with YOLOV4, $AP_{L}$ increased by 3.9. The reason why the effect on the large object is weaker than DERT is that this article guesses that compared to the loss of DETR, the improved loss of this article adds L2 loss and area error, which limits the size of the prediction frame. , Resulting in the model may be more inclined to those medium-scale prediction boxes. In the case where the improvement of $AP_{L}$ is low, this also increases the detection effect of $AP_{S}$ and $AP_{L}$, which can be considered as a relatively powerful size limit.  

In addition, the prediction results after pre-classification have been improved to a certain extent. We hypothesize that the calculation of confidence loss and matching loss added to the loss can penalize mismatches and mismatches to a greater extent than directly using IOU for loss calculations.

\begin{table}[h!]
  \begin{center}
    \caption{results}
    \label{table3}
    \begin{tabular}{lllllll} 
      \textbf{Model} & \textbf{$mAP$} & \textbf{$AP_{50}$} & \textbf{$AP_{75}$} & \textbf{$AP_{S}$} & \textbf{$AP_{M}$} & \textbf{$AP_{L}$}\\
      \hline
      DETR & 44.9 & 64.7 & 47.7 & 23.7 & 49.5 & \textbf{62.3}\\
      \hline
      YOLOV4 & 43.5 & 65.7 & 47.3 & 26.7 & 46.7 & 53.3\\
      \hline
      ESOD & \textbf{45.8} & \textbf{66.0} & \textbf{51.7} & \textbf{29.6} & \textbf{52.2} & 57.2\\
    \end{tabular}
  \end{center}
\end{table}

\subsection{costs}
The mAP, average inference energy consumption, and average inference time when only using the pre-classification model for scheduling on the local mobile platform are shown in the picture \ref{fig:result:cost}. At this time, compared to when only using DETR on the local platform for target detection tasks, mAP increases by \textbf{0.9}, the average inference time is reduced by \textbf{22.13}\%, and the average inference energy consumption is reduced by \textbf{29.6}\%.

\begin{figure}[htbp]
	\centering
	\includegraphics[scale=0.2]{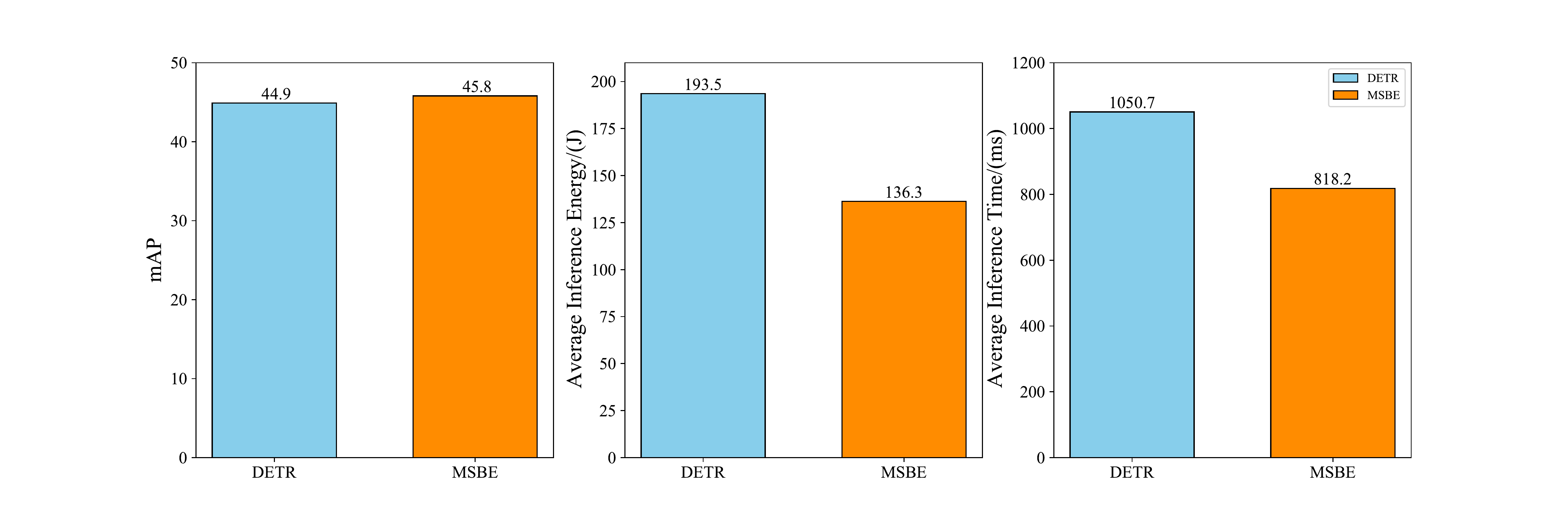}
	\caption{Lower overhead and higher MAP with ESOD compared with the SOTA model}
	\label{fig:acc:pre-model}
\end{figure}

The average inference energy consumption and average inference time when the edge server is added to offload computationally intensive tasks are shown in the table \ref{table5}. When time is used as the optimization criterion, that is, when the weight of time is increased to the maximum, the average inference time is reduced by 99.87\%, and average inference energy consumption by 95.5\%; when the weight of energy consumption is increased to the maximum, the average inference time is reduced by 99.22\%, and the average inference energy consumption is reduced by 98.04\%; set the weight of energy consumption and time When equal, the average inference time is reduced by 96.95\%, and the average inference energy consumption is reduced by 99.6\%.

\begin{table}[h!]
  \begin{center}
    \caption{strategy}
    \label{table5}
    \begin{tabular}{lllllll} 
      \textbf{Model} & \textbf{Infer Energy(J)} & \textbf{Infer Time(s)}\\
      \hline
      Balance Natively & 136.263 & 818.165 \\
      Time-Oriented& 6.197 & 1.025 \\
      Energy-Oriented & 2.672 & 6.389 \\
      Balance with Serve & 4.156 & 3.232 \\
    \end{tabular}
  \end{center}
\end{table}

\section{conclusion}

Aiming at the problem of the high computational cost of target detection algorithms on mobile devices, this paper proposes a target detection scheduling scheme based on edge devices, which improves the precision of target detection by using pre-classification steps with minimal cost. After the task is offloaded in conjunction with the edge server, compared to directly performing the target detection task on the mobile device, the energy consumption and time of the calculation can be significantly reduced. At the same time, the scheduling algorithm in this paper can switch between different indicators to achieve a balance between time and energy consumption.

\section*{Acknowledgment}
This work was Supported by National Key R\&D Program of China(No: 2019YFC1521400), the National Natural Science Foundation of China (No. 61902229), Fundamental Research Funds for the Central Universities (No. GK202103084) .

\end{document}